\documentclass[runningheads]{llncs}

\usepackage{graphicx}
\usepackage{wrapfig}
\usepackage{tabularx}
\usepackage{arydshln}
\usepackage{subcaption}
\usepackage[export]{adjustbox}
\usepackage{multirow}
\usepackage{dsfont}
\usepackage{xcolor,colortbl}
\usepackage{amsmath,amsfonts,bm,mathtools, bbm}
\usepackage{amssymb}
\usepackage{cite}
\usepackage{hyperref}
\usepackage{amssymb}
\usepackage{enumitem}
\usepackage{multirow}
\usepackage{tabularray}
\usepackage{booktabs}

\usepackage{pifont}

\usepackage{color}

\newcommand{\xmark}{\ding{55}}

\title{Few-shot Adaptation of \\ Medical Vision-Language Models}
\titlerunning{Few-shot Adaptation of Medical Vision-Language Models}

\author{Fereshteh Shakeri\inst{1,2}\thanks{Equal contributions.} \and
Yunshi Huang\inst{1,2}$^*$ \and Julio Silva-Rodríguez\inst{1} \and \\
Houda Bahig\inst{2} \and An Tang\inst{2} \and Jose Dolz\inst{1} \and Ismail Ben Ayed\inst{1,2}}
\authorrunning{Shakeri \textit{et al.}}
\institute{\'ETS Montreal \and Centre de Recherche du Centre Hospitalier de l’Université de Montréal (CRCHUM)}

\begin{document}

\maketitle              
\begin{abstract}
Integrating image and text data through multi-modal learning has emerged as a new approach in medical imaging research, following its successful deployment in computer vision. While considerable efforts have been dedicated to establishing medical foundation models and their zero-shot transfer to downstream tasks, the popular few-shot setting remains relatively unexplored. Following on from the currently strong emergence of this setting in computer vision, we introduce the first structured benchmark for adapting medical vision-language models (VLMs) in a strict few-shot regime and investigate various adaptation strategies commonly used in the context of natural images. Furthermore, we evaluate a simple generalization of the linear-probe adaptation baseline, which seeks an optimal blending of the visual prototypes and text embeddings via learnable class-wise multipliers. Surprisingly, such a text-informed linear probe yields competitive performances in comparison to convoluted prompt-learning and adapter-based strategies, while running considerably faster and accommodating the black-box setting. Our extensive experiments span three different medical modalities and specialized foundation models, nine downstream tasks, and several state-of-the-art few-shot adaptation methods. We made our benchmark and code publicly available to trigger further developments in this emergent subject: \url{https://github.com/FereshteShakeri/few-shot-MedVLMs} \ .

\end{abstract}

\section{Introduction}
\label{sec:intro}

Deep neural networks have attracted paramount attention in the last decade in the medical image analysis community \cite{reviwmedia}. Their breakthrough developments in natural image recognition tasks have been successfully applied to a breadth of medical tasks, such as radiology image classification \cite{irvin2019chexpert}, tumor grading in gigapixel stained histology images \cite{silva2020going}, or diabetic retinopathy grading \cite{messidor}, among others. However, the limitations of such models have restricted their widespread adoption in real clinical settings. In particular, they require large labeled datasets for training reliable task-specific solutions, a burden for medical domains \cite{Chen2022}, in which annotated data is usually scarce. In addition, the large domain drifts existing in medical image analysis from inter-scanner, inter-stain, or inter-population variability require continuous adaptation, ideally done in a data-efficient way, \textit{i.e.} using small numbers of labeled samples, a.k.a {\em few-shot} adaptation. A potential alternative for such adaptation is transfer learning of large pre-trained models that extract robust features. Although popular in computer vision, transferring such models from natural to medical images did not achieve the expected gains \cite{transfusion}, due to the fine-grained nature of medical images.

A paradigm shift in transfer learning is currently underway, focused on large-scale pre-training on heterogeneous datasets, which have shown improved transferability, the so-called \textit{foundation models}. In particular, vision-language models, such as CLIP \cite{radford2021clip} and ALIGN \cite{jia2021scaling}, exhibit remarkable adaptability to various downstream tasks. These models can integrate large-scale sources with text supervision (\textit{e.g.} 400M image-text pairs for CLIP), and train joint embedding representations of such modalities by contrastive learning, which have shown astonishing robustness to domain drifts \cite{radford2021clip}. In addition, such pre-trained knowledge can be efficiently transferred to downstream tasks, in low-shot regimes. Although those are conditions largely desired in the medical-imaging community \cite{Moor2023}, the direct application of CLIP models has been limited in this domain, since they lack fine-grained expert's medical knowledge. 

To alleviate this issue, a myriad of recent works have gathered large open-access medical datasets to build specialized medical vision-language models for radiology \cite{convirt,MedCLIP,medklip}, histology \cite{plip,ikezogwo2023quilt}, or ophthalmology \cite{FLAIR2023}. With the current endeavors towards developing and adapting such models to downstream tasks, nevertheless, there are important specific challenges inherent to clinical domains, which are largely being overlooked. First, current studies on medical VLMs predominantly revolve around fine-tuning models with a reduced percentage of the available datasets (e.g., $1\%$ or $10\%$ in \cite{ikezogwo2023quilt} or \cite{convirt}), which still amount to hundreds or thousands of annotated samples. This assumes large labeled datasets for adaptation, which might be inconvenient in clinical applications, particularly when dealing with rare, low-prevalence diseases. Second, pre-training medical foundation models will potentially involve the use of private sources of clinical records, both images and text reports. While recent studies have warned about the potential leaking of the source data from solely using the pre-trained weights \cite{leakdata}, fine-tuning the entire encoders during adaptation is still a dominant choice in the literature \cite{medklip}. Moreover, foundation models tend to improve performances by increasing substantially the number of trainable parameters, thereby requiring substantial hardware requirements for full fine-tuning, which may be unpractical in clinical institutions, with limited computational sources.

Linear probing (LP) is a standard adaptation method, which was also evaluated in the seminal CLIP paper \cite{radford2021clip}. It is a computationally efficient fine-tuning baseline, which operates in black-box settings, {\em i.e.} it does not require access to the inner representations of the pre-training models. It consists of updating the weights of a linear classifier on top of the frozen vision encoder, by optimizing the cross-entropy loss built with a few labeled samples in the target task. Unfortunately, LP has often been reported as a very weak baseline in the recent literature on few-shot VLMs \cite{radford2021clip,Zhou2022coop,zhang2022tip}, as it completely omits the text encoder's knowledge, potentially over-fitting the few labeled images. This has motivated intensive recent research efforts in computer vision, targeted at building convoluted {\em prompt learning} \cite{Zhou2022coop,Yao2023kgcoop,zhou2022cocoop,chen2023plot} or {\em feature-adaptation} \cite{Gao2023clipadapter,zhang2022tip} strategies, which account for such information. In particular, prompt learning is gaining wide popularity in the field. This parameter-efficient family of methods improves adaptation by optimizing the best text input for a target task, via learnable continuous prompts. We demonstrate that such prompt-tuning approaches offer limited performance gains in few-shot medical-image classification, at the cost of imposing an overlooked extensive computational and memory overhead, requiring gradient back-propagation throughout the entire text encoder. Moreover, the assumption of accessing the learned parameters of the text encoder may hinder their deployment in low-resource and privacy-preserving black-box scenarios, which are 
crucial considerations in medical domains. To address these issues, a few, very recent studies in computer vision have incorporated knowledge from the text encoder to enhance the linear-probe baseline \cite{lp24, lin2023multimodality}. 

Given the continuous emergence of foundation models in medical imaging, along with the potential deployment of these popular adaptation methods, we aim at paving the way towards more realistic adaptation of medical VLMs, taking into account transferability scenarios with access to limited labeled examples per task, {\em i.e.} up to 16 shots. Our main contributions could be summarized as follows: 
\begin{itemize}
\item We introduce the first structured benchmark for adapting medical vision-language models (VLMs) in a strict few-shot regime.
\item We evaluate a simple generalization of the LP baseline, which seeks an optimal blending of the visual prototypes and text embeddings via learnable 
class-wise multipliers. Surprisingly, such a text-informed LP yields competitive performances in comparison to convoluted prompt-learning and 
adapter-based strategies, while running considerably faster and accommodating the black-box setting (as it requires access to the output embeddings only).
\item We report extensive evaluations and comparisons over three different medical modalities and specialized foundation models, nine downstream tasks and several state-of-the-art 
few-shot adaptation strategies. 
\end{itemize}

\section{Related Work}
\label{sec:rw}

\noindent \textbf{\textit{Prompt learning.}} One of the foremost categories of approaches in the few-shot adaptation of vision-language models is prompt learning, motivated by the observation that the choice of input prompt may affect the performance of zero-shot prediction. Following the burgeoning interest in prompt learning within the NLP community \cite{ShinEMNLP2020,JiangACL2020,Hong2021NAACL}, notable work by \cite{Zhou2022coop} introduced context optimization (CoOp) for vision-language models. In CoOp, text is represented as learnable continuous vectors, which are trained as task-specific prompts through few-shot training examples and a standard supervised classification loss. 
The innovative idea of CoOp has spurred an extensive body of literature on prompt learning for few-shot vision-language models, yielding numerous sophisticated extensions \cite{zhou2022cocoop, chen2023plot, Yao2023kgcoop, Zhu2023prograd}. For example, CoCoOp \cite{zhou2022cocoop} additionally learns instance-conditional contexts conditioned on the inputs to improve the generalization of CoOp to unseen classes. PLOT \cite{chen2023plot} learns multiple prompts to describe each class's characteristics through the minimization of an optimal-transport distance. KgCoOp \cite{Yao2023kgcoop} enhances CoOp's performance on unseen classes by minimizing the discrepancy between the text embeddings generated by the learned prompts and hand-crafted ones. ProGrad \cite{Zhu2023prograd} aligns few-shot downstream knowledge with large-scale general knowledge, thereby mitigating overfitting the few-shot samples. Given the popularity of prompt-learning methods in vision and NLP, there is currently an emergent interest in their application within the medical field. This includes, for instance, parameter-efficient medical image segmentation \cite{fischer2024prompt} and prompt learning on large clinical language models \cite{taylor2023clinical}, the latter being more closely related to our setting.

\noindent \textbf{\textit{Black-box Adapters.}} Adapters represent another category of approaches within the realm of few-shot adaptation for VLMs. These methods focus on non-linear transformations applied to the pre-trained vision and text features \cite{Gao2023clipadapter,zhang2022tip}. They are multi-layer modules added to the encoder's bottleneck, and whose parameters are fine-tuned over a few-shot task by optimizing the cross-entropy loss. For instance, CLIP-Adapter \cite{Gao2023clipadapter} incorporated a multi-layer perceptron to learn new features, which are blended with the original pre-trained features through residual connections. Tip-Adapter \cite{zhang2022tip} integrated a non-linear, quadratic-complexity module to assess pairwise similarities between the features of the labeled samples, and blended the resulting class scores with the textual features. This category of approaches effectively alleviates the limitation of prompt-learning methods in terms of computational complexity, by eliminating the need for back-propagation over the text encoder. However, their performance relies heavily on key hyper-parameters, particularly those governing the blending between vision and textual features, which require computationally intensive grid searches.

\section{Methods}
\label{method}

\noindent \textbf{\textit{The few-shot image classification setting.}} Following on from the popular few-shot setting in computer vision \cite{Zhou2022coop,zhang2022tip}, our approach involves a foundation model pre-trained on a large dataset composed of image-text pairs. The objective is to predict the labels of samples from previously unseen target datasets, via fine-tuning on a limited number of labeled samples, a.k.a {\em the support set}. 
For each support image $\bm{x}_i$, one may compute its vision embedding $\bm{f}_{i}=\bm{\theta}_{v}(\bm{x}_{i})$, with $\bm{\theta}_{v}$ denoting the frozen pre-trained visual encoder. 
Also, for each given target class $k \in 1, \dots, K$, one may use a textual description of the class (or a prompt), $\bm{z}_{k}$, e.g., ``an image of a [class$_{k}$]'', where [class$_{k}$] is 
the class name/description. Let $\bm{t}_{k} = \bm{\theta}_{t}(\bm{z}_{k})$ denotes the corresponding text embedding, and $\bm{\theta}_{t}$ the text encoder.

\noindent \textbf{\textit{The standard linear-probe (LP) baseline.}} The standard linear probe (LP), initially evaluated as a few-shot adaptation baseline in the CLIP paper \cite{radford2021clip}, is a linear classifier that exclusively utilizes the frozen vision features. It optimizes the following cross-entropy loss w.r.t. the last-layer weights of the vision encoder (i.e., the class prototypes), $\mathbf{w} = (\bm{w}_{k})_{1 \leq k \leq K}$:
\begin{equation}
\label{CE-loss-lp}
    L_{\textrm{CE}} (\mathbf{w})= -\frac{1}{N}\sum_{i=1}^{N}\sum_{k=1}^{K} y_{ik}\ln{p_{ik}(\mathbf{w}}); \quad p_{ik}(\mathbf{w}) =   \frac{\exp \left ( \bm{f}_{i}^{t} \bm{w}_{k} \right) }{\sum_{j=1}^{K}\exp \left (\bm{f}_{i}^{t} \bm{w}_{j} \right )} 
\end{equation}
where $y_{ik}$ denotes one-hot encoded label of support image $\bm{x}_{i}$, i.e., $y_{ik}=1$ if $\bm{x}_{i}$ belongs to class $k$ and 0 otherwise.
Unlike prompt learning methods and Adapters, which integrate text knowledge, a limitation of this standard LP baseline is that it omits completely information from the text encoder, {\em i.e.} $\mathbf{t} = (\bm{t}_{k})_{1 \leq k \leq K}$, yielding significantly lower performances than zero-shot predictions \cite{radford2021clip}.

\noindent \textbf{\textit{Text-driven linear probe (LP+text).}} We evaluate a simple generalization of the LP baseline, which we introduced recently in the context of natural-image few-shot tasks \cite{lp24}. Our method integrates text knowledge while accommodating the black-box setting. It seeks an optimal blending of the visual prototypes and text embeddings via learnable class-wise multipliers, $\bm{\alpha} = (\alpha_k)_{1 \leq k \leq K}$, by optimizing the following loss function:
\begin{equation}
\label{CE-loss}
    L_{\textrm{CE}} (\mathbf{w}, \bm{\alpha} )= -\frac{1}{N}\sum_{i=1}^{N}\sum_{k=1}^{K} y_{ik}\ln{p_{ik}(\mathbf{w}, \bm{\alpha})}; \quad p_{ik}(\mathbf{w}, \bm{\alpha}) =   \frac{\exp \left ( \bm{f}_{i}^{t} (\bm{w}_{k} + \alpha_k \bm{t}_{k}) \right) }{\sum_{j=1}^{K}\exp \left (\bm{f}_{i}^{t} (\bm{w}_{j} + \alpha_j \bm{t}_{j}) \right )}  
\end{equation}

\noindent During few-shot adaptation, visual class prototypes $\mathbf{w} = (\bm{w}_{k})_{1 \leq k \leq K}$ and class-wise blending parameters $\bm{\alpha} = (\alpha_k)_{1 \leq k \leq K}$ are updated via full-batch gradient descent, while text embeddings $\mathbf{t} = (\bm{t}_{k})_{1 \leq k \leq K}$ are kept fixed. To minimize \eqref{CE-loss}, we follow the computationally efficient, full-batch optimizer in \cite{lp24}, in which step sizes are implicit (derived from the 
Lipschitz-gradient properties of the objective function \cite{lp24}). This relaxes intensive validation searches for the optimization hyper-parameters, unlike standard gradient descent practices where learning rates are intensively searched over validation sets. Therefore, it runs significantly faster than state-of-the-art few-shot adaptation methods for VLMs.

\section{Experiments}
\label{sec:experiments}

\noindent \textbf{\textit{Medical vision-language models (VLMs).}} A comprehensive assessment of the potential of medical VLM adaptation is carried out across three different popular medical domains: histology, radiology, and ophthalmology. In each domain, we utilize an open-access specialized foundation VLM. \textbf{Histology}: we employed Quilt-1M \cite{ikezogwo2023quilt}, with ViT-B/32 vision and GPT2 text encoder. \textbf{Ophtalmology}: we utilized FLAIR \cite{FLAIR2023}, a foundation model focused on color fundus image understanding. \textbf{Radiology}: we focused on chest X-ray (CXR) scans, which have attracted the attention of a large body of literature \cite{convirt,MedCLIP,medklip}. Concretely, we used MedCLIP \cite{MedCLIP} pre-trained on CheXpert \cite{irvin2019chexpert} and MIMIC-CXR \cite{johnson2019mimic} datasets. Since these datasets are also further used for evaluation, we pre-trained this model to control test partition better and avoid test-data leakage. We followed \cite{MedCLIP} implementation details. Note that FLAIR and MedCLIP present a similar dual-encoder architecture: ResNet-50 as vision encoder, and BioClinicalBERT \cite{bioclinicalbert} text encoder. It is worth mentioning that those models cover a wide range of architectures, both convolutional and ViTs.

\noindent \textbf{\textit{Adaptation tasks.}} Our benchmark encompasses a wide number of downstream tasks for the adaptation of medical VLMs. To ensure a logical transfer of the pre-trained features, each specialized foundation model is used uniquely for datasets from their respective domain. In addition, such open-access datasets are carefully selected to avoid test data leaking, \textit{i.e.} evaluating with data used for pre-training. \textbf{Histology}: involve three different organs and cancer types. Concretely, colorectal adenocarcinoma samples in NCT-CRC \cite{kather2018100}, prostate cancer grading in SICAPv2 \cite{silva2020going}, and SkinCancer \cite{kriegsmann2022deep}. \textbf{Ophtalmology}: we consider MESSIDOR \cite{messidor} focused on diabetic retinopathy (DR) grading, and FIVES \cite{fives} and ODIR200x3 \cite{odir}, for inter-diseases discrimination. \textbf{Radiology}: following the same evaluation benchmark as in \cite{MedCLIP}, we employed CheXpert$_{5\times200}$\cite{irvin2019chexpert}, MIMIC$_{5\times200}$\cite{johnson2019mimic}, and RSNA \cite{rsna}. These datasets include a heterogeneous variety of fine-grained findings, such as pneumonia, atelectasis, edema, or pleural effusion.

\noindent \textbf{\textit{Few-shot adaptation protocol and evaluation.}} Transfer learning from the large-scale pre-trained models is performed in a challenging, but realistic medical setting, in which only a few samples, \textit{i.e.} shots, are available. Following relevant literature in natural image \cite{radford2021clip,Zhou2022coop,Gao2023clipadapter}, the training subset consists of $S= \{1,2,4,8,16\}$ images per class randomly sampled for each dataset in all scenarios. To guarantee fair comparisons among different approaches, we deploy a few-shot validation set with the same number of samples for hyper-parameters tuning. We employed the test splits from the original datasets, if available, or performed a $20\%$ hold-out partition otherwise. The evaluation metric is a balanced average accuracy (ACA), widely employed in CXR \cite{MedCLIP} and Ophthalmology \cite{FLAIR2023} benchmarks. The evaluation is carried out through $5$ random seeds to account for the variability in the few shots selected.

\noindent \textbf{\textit{Implementation details and baselines.}} We conduct a comprehensive comparison of several state-of-the-art methods in the few-shot efficient transfer learning of CLIP-based models. Our benchmarks include Zero-shot prediction (\textit{i.e.} no adaptation), Prompt Learning, and black-box Adapter methods. \textbf{Zero-shot}: following CLIP \cite{radford2021clip}, these predictions are obtained by computing the softmax cosine similarity between image and text embeddings. Text embeddings for each category are obtained following the specific prompts used in each original VLM's publication. This is, prompt ensembles for MedCLIP \cite{MedCLIP}, and domain-expert descriptions for FLAIR \cite{FLAIR2023} and Quilt-1M \cite{ikezogwo2023quilt}. It is worth mentioning that the same text-driven prompts are used when required in other Adapters. \textbf{Prompt Learning}: we resort to the popular CoOp \cite{Zhou2022coop} and KgCoOp \cite{Yao2023kgcoop}. \textbf{Black-box Adapters}: The firstly proposed linear probing in CLIP paper, LP, is considered as a baseline. Concretely, logistic regression is trained with the L-BFGS \cite{nocedal1980updating} optimizer. Also, more recent adaptation techniques such as CLIP-adapter \cite{Gao2023clipadapter} and TIP-adapter \cite{zhang2022tip} are included. For TIP-Adapter, we employed its fine-tuned version, TIP-Adapter-F, and set $\alpha$ and $\beta$ to 1 initially. Later, we find best values of $\alpha$ and $\beta$ using the validation set. Finally, we include the efficient proposed LP+text in the benchmark, following its description in Section \ref{method}.

\begin{figure}[h!]
    \centering
    \includegraphics[width=\linewidth]{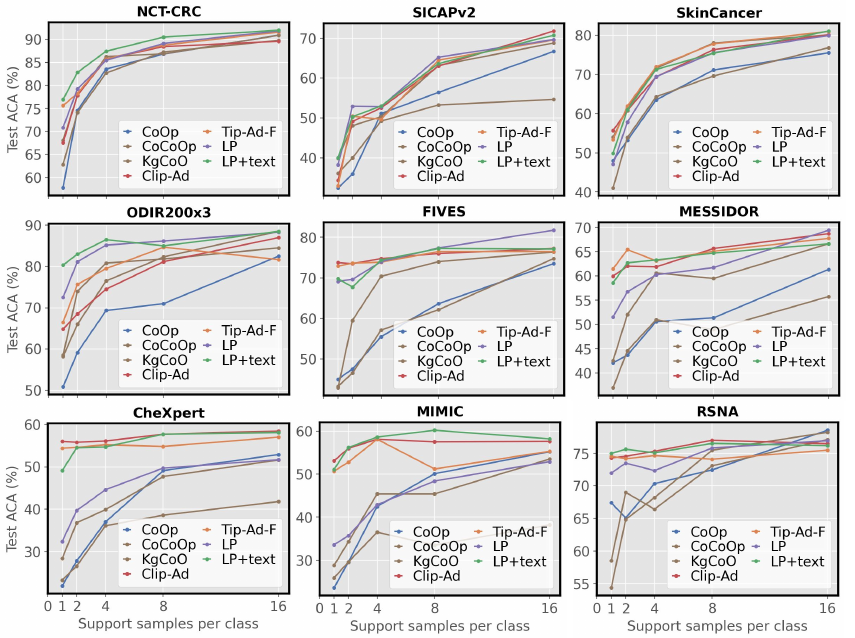}
    \caption{Comparison of different adaptation methods of Medical VLMs evaluated on 9 benchmarks, averaged over 5 tasks. 
    } 
    \label{fig:main}
\end{figure}

\noindent \textbf{\textit{Results.}} Figure \ref{fig:main} shows a quantitative comparison of all studied few-shot adaptation methods on the 9 benchmarks. As demonstrated by the figure LP+text performs relatively well in most cases, outperforming prompt learning methods by a large margin and performing on par with Adapters. In Table~\ref{table:avg-main} we present specific numerical results for each method, averaged per modality. Specific numeric results per dataset are provided in Supp. Materials. It is worth mentioning from the results that Prompt Learning methods rarely outperform black-box Adapters. For instance, the most recent method of such a family, KgCoOp \cite{Yao2023kgcoop} ranges performance drops (\textit{e.g.} $[1.3, 3.4]\%$ for S=16) compared with the proposed LP+text. In addition, the significant standard derivation of prompt learning is relatively large, especially in low-shot settings, which motivates the use of Adapters as a more appealing alternative. Comparing the proposed LP+text with other Adapters, our method shows consistent performance gains to the popular TIP-Adapter \cite{zhang2022tip}, and performs at par with CLIP-Adapter, albeit being much more computationally efficient, as we later discuss. Finally, while the basic LP suffers a consistent performance drop in the extreme-low data regime (\textit{i.e.} S=1), introducing text information in LP+text prevents it.

\begin{table}[h!]
  \scriptsize
    \caption{\textbf{Comparison of state-of-the-art methods.} Average ACA (\%) on 3 benchmarks for each modality. Best values are highlighted in \textbf{bold}.} \label{table:avg-main}
    \resizebox{1\textwidth}{!}{
    \begin{tabular}{lccccc}
        \toprule
        \textbf{(a) Histology} & \textbf{$S$=1} & \textbf{$S$=2} & \textbf{$S$=4} & \textbf{$S$=8} & \textbf{$S$=16}\\
        \midrule
        Zero-shot $_\text{ ICML'21}$\cite{radford2021clip} & \multicolumn{5}{c}{48.33}\\
        \midrule
        CoOp$_\text{ IJCV'22}$\cite{Zhou2022coop} & $46.05 \pm 9.79$ & $54.55 \pm 8.53$  &  $66.04 \pm 5.15$  & $71.45 \pm 5.53$  &  $77.69 \pm 1.32$ \\
        CoCoOp$_\text{ CVPR'22}$\cite{chen2023plot} & $46.63 \pm 7.71$ &  $55.98 \pm 5.65$ &  $65.39 \pm 3.04$  & $70.04 \pm 3.05$  & $73.73 \pm 2.83$  \\
        KgCoOp$_\text{ CVPR'23}$\cite{Yao2023kgcoop} & $53.96 \pm 5.95$ & $62.44 \pm 3.49$ & $69.37 \pm 3.33$ &$76.01 \pm 2.46$ & $79.91 \pm 1.07$ \\
        \midrule
        CLIP-Adapter$_\text{ IJCV'23}$\cite{Gao2023clipadapter} & $52.50 \pm 8.31 $ & $62.58 \pm 4.07$ & $69.21 \pm 4.44$ & $75.92 \pm 2.48$ & $ 80.47 \pm 1.31$ \\
        Tip-Adapter-F$_\text{ ECCV'22}$\cite{zhang2022tip} & $53.97 \pm6.11$ & $63.54 \pm3.41$ & $69.11 \pm4.24$  & $\textbf{77.01}\pm 2.52$  & $80.69\pm 1.42$ \\
        \midrule
        Linear probe (LP) & $52.05 \pm 4.66$  &  $63.33 \pm 3.24$  &  $69.22 \pm4.02$  &  $76.64 \pm1.66$  & $80.47\pm 1.61$  \\
        LP+text \cite{lp24} & $\textbf{55.60} \pm 6.26$ & $\textbf{64.69} \pm 3.65$ & $\textbf{70.56} \pm 3.94$ & $76.52 \pm 2.44$ & $\textbf{81.26} \pm 1.76$\\
        \bottomrule 
        \\
    
        \toprule
        \textbf{(b) Ophtalmology} & \textbf{$S$=1} & \textbf{$S$=2} & \textbf{$S$=4} & \textbf{$S$=8} & \textbf{$S$=16}\\
        \midrule
        Zero-shot $_\text{ ICML'21}$\cite{radford2021clip} & \multicolumn{5}{c}{65.74}\\
        \midrule
        CoOp$_\text{ IJCV'22}$\cite{Zhou2022coop} & $45.98 \pm 12.26$ & $50.11 \pm 12.29$  &  $58.48 \pm 11.12$  & $62.00 \pm 6.96$  &  $72.45 \pm 2.04$ \\
        CoCoOp$_\text{ CVPR'22}$\cite{chen2023plot} & $47.87 \pm 12.07$ &  $59.19 \pm 7.97$ &  $69.16 \pm 5.79$  &  $71.94 \pm 4.43$ & $77.16 \pm 3.01$  \\
        KgCoOp$_\text{ CVPR'23}$\cite{Yao2023kgcoop} & $46.23 \pm 10.26$ & $55.03 \pm 8.01$ & $62.98 \pm 4.49$ &$64.31 \pm 4.92$ & $71.67 \pm 4.98$ \\
       \midrule
        CLIP-Adapter$_\text{ IJCV'23}$\cite{Gao2023clipadapter} & $66.18 \pm 4.54 $ & $68.00 \pm 4.29$ & $70.38 \pm 5.90$ & $74.27 \pm 3.99$ & $ 77.65 \pm 2.72$ \\
         Tip-Adapter-F$_\text{ ECCV'22}$\cite{zhang2022tip} & $66.95 \pm4.03$ & $\textbf{71.57} \pm3.78$ & $72.16 \pm3.92$  & $75.42\pm 4.12$  & $75.30\pm 3.38$ \\
       \midrule
         Linear probe (LP) & $64.39 \pm 5.57$  &  $69.18 \pm 5.28$  &  $73.13 \pm4.38$  &  $75.09 \pm4.24$  & $\textbf{79.83}\pm 2.34$  \\
         \cellcolor{blue!15}{LP+text \cite{lp24}} & \cellcolor{blue!15}{$\textbf{69.56} \pm 6.22$} & \cellcolor{blue!15}{$71.15 \pm 4.95$} & \cellcolor{blue!15}{$\textbf{74.72} \pm 3.80$} & \cellcolor{blue!15}{$\textbf{75.66} \pm 3.42$} & \cellcolor{blue!15}{$77.42 \pm 2.07$}\\
        \bottomrule 
        \\
        
    \toprule
    \textbf{(c) Radiology} & \textbf{$S$=1} & \textbf{$S$=2} & \textbf{$S$=4} & \textbf{$S$=8} & \textbf{$S$=16}\\
    \midrule
    Zero-shot $_\text{ ICML'21}$\cite{radford2021clip} & \multicolumn{5}{c}{60.37}\\
    \midrule
    CoOp$_\text{ IJCV'22}$\cite{Zhou2022coop} & $37.64 \pm 6.82$ & $40.82 \pm 6.76$  &  $49.95 \pm 6.15$  & $57.21 \pm 3.97$  &  $62.21 \pm 4.00$ \\
    CoCoOp$_\text{ CVPR'22}$\cite{chen2023plot} & $34.52 \pm 6.50$ &  $40.35 \pm 5.63$ &  $46.93 \pm 6.60$  &  $49.19 \pm 4.55$ & $52.73 \pm 3.46$  \\
    KgCoOp$_\text{ CVPR'23}$\cite{Yao2023kgcoop} & $38.57 \pm 7.47$ & $46.70 \pm 7.11$ & $50.57 \pm 5.72$ &$55.39 \pm 3.47$ & $60.73 \pm 3.51$ \\
    \midrule
    CLIP-Adapter$_\text{ IJCV'23}$\cite{Gao2023clipadapter} & $\textbf{61.13} \pm 2.43 $ & $\textbf{62.10} \pm 2.66$ & $\textbf{63.17} \pm 2.93$ & $64.06 \pm 2.48$ & $ \textbf{64.15} \pm 2.27$ \\
    Tip-Adapter-F$_\text{ ECCV'22}$\cite{zhang2022tip} & $59.88 \pm2.80$ & $60.52 \pm1.68$ & $62.64 \pm4.55$  & $60.03\pm 3.29$  & $62.59\pm 2.47$ \\
    \midrule
    Linear probe (LP) & $45.98 \pm 4.87$  &  $49.63 \pm 4.50$  &  $53.28 \pm4.80$  &  $57.97 \pm3.12$  & $60.50\pm 4.76$  \\
    \cellcolor{blue!15}{LP+text \cite{lp24}} & \cellcolor{blue!15}{$58.39 \pm 5.03$} & \cellcolor{blue!15}{$\textbf{62.10} \pm 3.80$} & \cellcolor{blue!15}{$62.79 \pm 3.19$} & \cellcolor{blue!15}{$\textbf{64.80} \pm 2.79$} & \cellcolor{blue!15}{$\textbf{64.15} \pm 3.20$}\\  
    \bottomrule
\end{tabular}
 }
\end{table}

\begin{table}[h!]
\scriptsize
\setlength{\tabcolsep}{4pt}
\caption{\textbf{Computational Efficiency.} Experiments on a single NVIDIA RTX A6000 GPU on NCT-CRC. $D_{1}=256$, and $D_{2}=D=512$. Number of context tokens for CoOp and KgCoOp: $n_{ctx1} = 16$; for CoCoOp: $n_{ctx2}=4$.
}
\label{table:time}
\centering
\begin{tabular}{lcrcc}
\toprule
 Methods & Category & \multicolumn{1}{c}{Training Time} & Black-box & \#Parameters\\
\midrule
Zero-shot\cite{radford2021clip} & & n/a & $\checkmark$ & n/a \\
\midrule
CoOp\cite{Zhou2022coop} & \multirow{3}{*}{\textit{Prompt-Learning}} & 3min & \xmark & $K\times n_{ctx1} \times D$\\
CoCoOp\cite{zhou2022cocoop} & & 12min& \xmark & $n_{ctx2}\times D+C$ \\
KgCoOp\cite{Yao2023kgcoop} & & 3min& \xmark & $K\times n_{ctx1} \times D$ \\
\midrule
 Clip-Adapter\cite{Gao2023clipadapter} & \multirow{2}{*}{\textit{CLIP-based Adapters}} & 2min & $\checkmark$ & $2(D_{1}\times D_{2})$\\
Tip-adapter-F\cite{zhang2022tip} & & 2min & $\checkmark$ & $K\times S\times D$ \\
\midrule
LP & \multirow{2}{*}{\textit{Linear probe}} & 43s & $\checkmark$ & $K \times D$\\
LP+text \cite{lp24} & &4s &$\checkmark$ & $K(D+1)$\\
\bottomrule
\end{tabular}
\end{table}

\noindent \textbf{\textit{Assessing computational workload.}} Here we evaluate the efficiency of the methods considered by presenting their computational overhead. We also indicate whether these methods enable black-box adaptation, which is a crucial consideration for addressing practical, real-world demands. Furthermore, we outline the number of parameters to be learned during training as an indicator of model complexity. This comparison Table~\ref{table:time} shows that, beyond outperforming state-of-the-art methods as shown in previous sections, LP+text stands out as the most efficient method. Complementary, it is worth noting that LP+text uses around 800MB of peak GPU memory, whereas CoCoOP requires up to 28GB (based on NCT-CRC experiments). This makes prompt learning methods inefficient for institutions with limited access to high-resource GPUs.

\paragraph{\textbf{Conclusions}.}
Inspired by the computer vision field, we established a new few-shot adaption setting for medical VLMs. We also introduced a generalization of LP baseline, integrating image and text embeddings through learnable class-wise multipliers. Evaluations across various benchmarks show that the proposed method stands out for its performance in different scenarios, its simplicity,  generalization ability, and its potential applicability in black-box scenarios.

\begin{credits}
\subsubsection{\ackname} This work was funded by the Natural Sciences and Engineering Research Council of Canada (NSERC) and Montreal University Hospital Research Center (CRCHUM). We also thank Calcul Quebec and Compute Canada.

\end{credits}

\bibliographystyle{splncs04}
\bibliography{biblio}

\clearpage

\end{document}